\documentclass{llncs}
\usepackage{makeidx}  
\usepackage{amsmath,amsfonts,amssymb}
\usepackage{graphicx}
\usepackage[misc]{ifsym}
\usepackage[tight]{subfigure}
\usepackage{caption} 
\newcommand{\vx}{\boldsymbol{x}}
\newcommand{\vmu}{\boldsymbol{\mu}}

\begin{document}
\frontmatter          
\pagestyle{headings}  
\mainmatter              
\title{Superpixel-Based Background Recovery from Multiple Images}
\titlerunning{Superpixel-Based Background Recovery}  
%
\author{Lei~Gao\inst{1}{(\Letter)} \and
Yixing Huang\inst{1} 
\and Andreas~Maier\inst{1,2}
}

\authorrunning{Lei Gao et al.} 

%
%
\institute{Pattern Recognition Lab, Friedrich-Alexander University Erlangen-Nuremberg, 91058 Erlangen, Germany 
\\
\email{lei.gao@fau.de} \\
\and
Erlangen Graduate School in Advanced Optical Technologies
(SAOT), 91058 Erlangen, Germany}

%
%
\maketitle              

\begin{abstract}
In this paper, we propose an intuitive method to recover background from multiple images. The implementation consists of three stages: model initialization, model update, and background output. We consider the pixels whose values change little in all input images as background seeds. Images are then segmented into superpixels with simple linear iterative clustering. When the number of pixels labelled as background in a superpixel is bigger than a predefined threshold, we label the superpixel as background to initialize the background candidate masks. Background candidate images are obtained from input raw images with the masks. Combining all candidate images, a background image is produced. The background candidate masks, candidate images, and the background image are then updated alternately until convergence. Finally, ghosting artifacts is removed with the k-nearest neighbour method. An experiment on an outdoor dataset demonstrates that the proposed algorithm can achieve promising results.

\end{abstract}
\section{Introduction}
In most popular tourism places, it is usually difficult for tourists to take a photo without other tourists walking around on camera. An optional choice for tourists is to recover a desired background scenery image from a series of images taken in burst mode. Background recovery is a basic task of background modelling \cite{BOUWMANS20173}.

So far, many background modelling methods have been proposed. Some methods presume that clean background images without foreground objects need to exist in the dataset \cite{BOUWMANS20173}. However, the assumption is not satisfied in the application. Therefore, in this work, we aim to recover a background scenery image from a series of images containing both background and foreground objects.

Depending on data abstraction, background modelling methods are classified into three categories \cite{BOUWMANS20173}: pixel-level \cite{TemporalMedian,784637,s120912279,NovelRobustStatistica,Automaticocclusion}, region-level \cite{spatiotemporalcontinuities,PatchBasedBackground,Backgroundrecovery,SuperpixelsMotion,SuperpixelsSimilarity}, and hybrid methods \cite{Extractingessence}.

Among pixel-level methods, Huang et al. \cite{TemporalMedian} have proposed a speed-up temporal median filter for background modelling. But it still requires that all background pixels exist in more than 50\% of frames. Gaussian mixture models (GMMs) \cite{784637} and kennel density estimation (KDE) methods \cite{s120912279} are often used to estimate the true probability density function of pixels value. But GMMs are difficult to determine the number of Gaussian distributions, while KDE is sensitive to a predefined band-width parameter. Both of them tend to misclassify foreground pixels, if a foreground pixel occurs more frequently than a background pixel at the same position.
Wang and Suter \cite{NovelRobustStatistica} chose the pixel value which varies little along the time-line as a background pixel, since the value of a background pixel remains fixed. Nevertheless, it fails when a foreground object moves slowly or stays stationary in several frames.
Herley et al. \cite{Automaticocclusion} assume that if a pixel has same value in any two images, it is a background pixel, which is a strong assumption.

Among region-level methods, Varadarajan et al. \cite{spatiotemporalcontinuities} employed change detection on a certain number of the beginning frames, categorizing pixels into background, weak foreground, and strong foreground according to two predefined thresholds. Each categorized pixel is modified recursively according to its 8-connected neighbourhood to remove outliers. Afterwards, spatial continuity and temporal continuity metrics are used to remove ghosting artifacts. However, the spatial continuity metric cannot handle complex situations and the temporal continuity metric cannot deal with temporarily static foreground objects.

Depending on the type of regions, region-level methods can be further categorized into patch-based methods and superpixel-based methods. A patch-based method proposed by Colombari and Fusiello \cite{PatchBasedBackground} can cope with heavy clutter.
Shrotre and Karam \cite{Backgroundrecovery} recover background by measuring a 5-D (L, a, b, color channel, and gradients in horizontal and vertical directions) patch-based distance metric. However, it may fail to recover a background patch if a foreground object stays stationary in several frames. In addition, all patch-based methods may  result in blocking artifacts and seams \cite{RandomWalk} .

Unlike patches split without specific meanings, superpixels \cite{superpixelsstateoftheart} are visually meaningful entities. 
Matrix decomposition into low-rank and sparse components is an effective method for background subtraction \cite{guyon2012robust}, but the computational cost is very high. Javed et al. \cite{SuperpixelMatrix} presented a superpixel-based background subtraction method to improve matrix decomposition performance.
Lim and Han \cite{SuperpixelsMotion} proposed to estimate motion models based on superpixels, and compute foreground and background probabilities for each pixel.
Fang et al. \cite{SuperpixelsSimilarity} found pixel value changes are more similar in a superpixel than their ordinary neighbours. This neighbourhood information is useful for background subtraction.

In this work, we propose a novel and intuitive hybrid method based on superpixels. It can overcome the drawbacks caused by patch-based methods mentioned above. The remainder of this paper is organized as follows. Section 2 describes the algorithm's details. The results on a real outdoor dataset are presented in Section 3. Section 4 concludes the paper.

\section{Method}

\begin{figure}
\centering
\includegraphics[width = \textwidth]{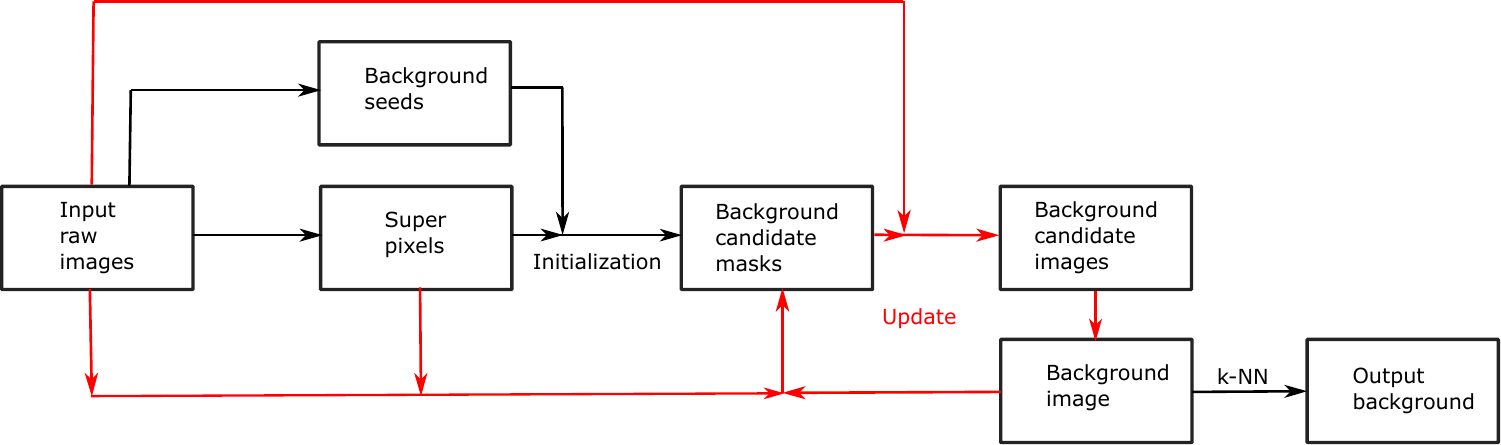}
\caption{The flowchart of the proposed background recovery method. The red paths update the background image until convergence.}
\label{Fig:flowchart}
\end{figure}

 The flowchart of our background recovery method is displayed in Fig.~\ref{Fig:flowchart}.

The idea of this background recovery method is inspired by \cite{PatchBasedBackground}, in which the most stable patches along time-line are chosen to start background tessellation. In our approach, only pixels vary little along the time-line are set as background seeds, which is a weaker assumption. These seeds then ``grow" with superpixels until convergence.

In this work, superpixels are created by simple linear iterative clustering (SLIC) \cite{SLIC}, as it is fast with a computational complexity of O(N). If the number of pixels labelled as background in a superpixel is above a predefined threshold, all pixels in the superpixel are labelled as background.

Background candidate masks are initialized for each input image. These Proper pixels are chosen in the candidate images to initialize a background image. Candidate masks are then updated by the background image. Afterwards candidate masks and the background image are updated alternatively until model convergence. In the end, artifacts in the background image are removed with the k-nearest neighbour (k-NN) method based on image spatial coherency \cite{Extractingessence}.

\subsection{Background seeds}
Firstly, a set of input RGB images are denoted by $\boldsymbol{I}={\{\boldsymbol{I}^1,...,\boldsymbol{I}^k\}}$ where $k$ is the total number of input images, and a pixel at (x,y) in the $i$-th image is denoted by $\boldsymbol{I}^i(x,y)$. We further denote a pixel value change image by $\boldsymbol{C}$, which is computed as the following,

 \begin{equation}
 \begin{array}{l}
 \boldsymbol{I}^c_{\max}(x,y)=\max\{..., {\boldsymbol{I}^{i,c}(x,y)}, ...\},\\

\boldsymbol{I}^c_{\min}(x,y)=\min\{...,{\boldsymbol{I}^{i,c}(x,y)},...\},\\

\boldsymbol{C}(x,y) = \max\{...,\boldsymbol{I}^c_{\max}(x,y) -  \boldsymbol{I}^c_{\min}(x,y),... \}, \\
 \end{array}
\end{equation}
for $c\in \{R,G,B\}$ and $i\in \{1,...,k\}$.
A seed image $\boldsymbol{I}_{\text{seed}}$ is computed from $\boldsymbol{C}$ with a predefined threshold $\tau_0$,
 \begin{equation}
\boldsymbol{I}_{\text{seed}}(x,y)
=\left\lbrace
\begin{array}{ll}
1, & \text{ if } \boldsymbol{C}(x,y)\leq \tau_0,\\
0, &\text{ if } \boldsymbol{C}(x,y)>\tau_0.
\end{array}
\right.
\label{eqn:seeds}
\end{equation}
  Because the seeds are the key points to successfully reconstruct the background image, the threshold should be carefully selected.  

\subsection{Superpixels}
In this work, superpixels are produced by SLIC. The general idea of SLIC is to split an image into grids and cluster pixels in neighbour grids according to a distance metric named 5-D distance ($D_s$) \cite{SLIC}. For initialization, each grid is a superpixel. Afterwards, the 5-D distance of each pixel to its neighbour superpixel centers is computed as follows,
 \begin{equation}
 \begin{array}{l}
 d_{l,a,b}=\sqrt{(l_k-l_i)^2+(a_k-a_i)^2+(b_k-b_i)^2},\\
d_{x,y}=\sqrt{(x_k-x_i)^2+(y_k-y_i)^2},\\
D_s=d_{l,a,b}+\frac{m}{s}d_{x,y},
 \end{array}
\end{equation}
where $l$, $a$ and $b$ are pixel values in CIELAB color space, $x$ and $y$ are the pixel coordinates, $s$ is grid interval distance, and $m$ is a weight parameter to adjust the compactness of a superpixel. Each pixel is clustered to its closest superpixel. This process updates superpixels iteratively until convergence.

\subsection{Background candidate mask initialization}

A binary background candidate mask is created for each input image. It is denoted by $\boldsymbol{I}_{\text{mask}}^i$ for the $i^\text{th}$ input image. The pixel values of each mask are set from the background seeds as
 \begin{equation}
\boldsymbol{I}^i_{\text{mask}}=\boldsymbol{I}_{\text{seed}}, \text{ for}\  i\in \{1,...,k\}.
\end{equation}

\textbf{The first rule for background candidate mask update}:
if the number of pixels labelled as background in a superpixel of an input image is above a predefined threshold $\tau_1$, or the ratio between the number of labelled pixels to the total number of pixels in a superpixel is above another predefined threshold $\tau_2$, all pixels in the superpixel are considered as background.We update the corresponding background candidate mask as follows,
 \begin{equation}
{\boldsymbol{I}^{i,j}_{\text{mask}}(x,y)}=1, \text{if }
 \#{\boldsymbol{I}^{i,j}_{\text{mask}}(x,y)}>\tau_1  \text{ or }
  \frac {\#\boldsymbol{I}^{i,j}_{\text{mask}}(x,y)}
          {\#\boldsymbol{I}^{i,j}(x,y)}>\tau_2
\label{eqn:tau1}
\end{equation}
 where $i\in \{1,...,k\}$, $\#\boldsymbol{I}^{i,j}_{\text{mask}}(x,y)$ is the number of pixels inside the $j^\text{th}$ superpixel in the $i^\text{th}$ image.

The update rule is based on background spatial continuity. Now the seeds are ``growing". 
\subsection{Background candidate images}
A background candidate image is created for each input image. It is denoted by $\boldsymbol{I}_{\text{cand}}^i$ for the $i^\text{th}$ input image. This candidate image is a pixel-wise operation between each channel of an input image and its corresponding mask as
 \begin{equation}
\boldsymbol{I}_{\text{cand}}^{i,c}= \boldsymbol{I}_{\text{mask}}^i\otimes\boldsymbol{I}^{i,c}, \text{ for }  i\in \{1,...,k\},
\end{equation}
where $\otimes$ stands for pixel-wise multiplication.

\subsection{Background initialization}
A background image $\boldsymbol{I}_{\text{back}}$ is initialized by choosing pixels from all candidate images. If a pixel is not labelled as background in all masks,  the corresponding pixel in $\boldsymbol{I}_{\text{back}}$ is set to zero. If a pixel is labelled only once in all masks, it is chosen for the corresponding pixel in $\boldsymbol{I}_{\text{back}}$. Otherwise, a pixel is chosen from the pixels by the mean shift method. This initialization process can be represented as the following,
 \begin{equation}
\boldsymbol{I}_{\text{back(x,y)}}=\left\lbrace
\begin{array}{lll}
\boldsymbol{I}^i_\text{cand}(x,y), & \text{ if }\#(\boldsymbol{I}^i_{\text{mask}}(x,y)=1 )=1,\\
\mathcal{M}\{...,\boldsymbol{I}^i_\text{cand}(x,y), ...\},  & \text{ if }\#(\boldsymbol{I}^i_{\text{mask}}(x,y)=1 )>1,\\
0,  &  \text{otherwise},
\end{array}
\right.
\end{equation}
where $i \in \{1,...,k\}$, $\mathcal{M}$ is the mean shift operation, and $\#(\boldsymbol{I}^i_{\text{mask}}(x,y)=1 )$ is the number of pixels whose value equals to 1 at $(x,y)$ in all  masks.

For each channel at each position $(x, y)$, the mean vector $\vmu$ is initialized as the mean of all the background candidate pixels whose corresponding mask pixels equal to 1. The mean vector is updated iteratively as follows \cite{meanshift},
 \begin{equation}
\vmu^{(l+1)}=\frac{\sum^n_{i=1} \vx_i g\left(\left |\left|\frac{\vmu^{(l)}-\vx_i}{h^{(l)}}\right|\right|^2\right)}{\sum^n_{i=1} g\left(\left |\left|\frac{\vmu^{(l)}-\vx_i}{h^{(l)}}\right|\right|^2\right)},
\end{equation}
where $l$ is the iteration number, $g(x)$ is a kernel function,
 \begin{equation}
g(x)=\left\lbrace
\begin{array}{ll}
1, & x\leq 1,\\
0, &x>1,
\end{array}
\right.
\end{equation}
 and $h$ is a scale normalization factor to represent the radius of the kernel. In this work, the radius $h$ is adaptively set to half of the difference between the maximum and the minimum elements within the previous radius, i.\,e.,
 \begin{equation}
 \begin{array}{l}
 h^{(l+1)} = (\max\{...,\vx_i,...\} - \min\{...,\vx_i,...\})/2,\\
 \text{ for } i \in \{i| g\left(\left |\left|\frac{\vmu^{(l)}-\vx_i}{h}\right|\right|^2\right)=1\},
 \end{array}
\end{equation}
 When there are only two pixels within the radius, the pixel value in the background image is set to  the first pixel's value.

\subsection{Model update}

Obviously, the number of pixels labelled as background pixels in the background image $\boldsymbol{I}_{\text{back}}$ is more than the numbers of pixels labelled as background in any masks. 

\textbf{The second rule for background candidate mask update}:
a pixel in the candidate mask is labelled as background if the difference of its corresponding raw input pixel value and the its corresponding background pixel value is smaller than a predefined threshold $\tau_3$,
 \begin{equation}
 \begin{array}{l}
 \boldsymbol{I}^i_{\text{mask}}(x,y)=1,
\text{ if }  |\boldsymbol{I}_{\text{back}}(x,y)-\boldsymbol{I}^i(x,y)|\leq \tau_3,
 \end{array}
 \label{eqn:tau3}
\end{equation}
for $i\in \{1,...,k\}$, and $(x,y)$ is labelled as background in $\boldsymbol{I}_{\text{back}}$.

Based on the above two rules, the background candidate masks $\boldsymbol{I}_{\text{mask}}$, background candidate images $\boldsymbol{I}_\text{cand}$, and the background image $\boldsymbol{I}_{\text{back}}$ are updated alternately until the background image $\boldsymbol{I}_{\text{back}}$ converges. 

\subsection{Output model}
After the above update, some ghosting artifacts may exist in the background image $\boldsymbol{I}_{\text{back}}$, which are caused by foreground object segmentation failure or pixels in candidate masks that cannot be updated based on the two rules. 

We notice that all ghosting artifact boundaries can be detected by the Canny edge detector. The edges and their neighborhood are considered as ghosting artifact candidates. An artifact candidate is then replaced by a pixel from an input image which is closest to the background image. The closeness is measured as Gaussian weighted sum of thresholded differences between neighbor pixels of the pixel and a patch of pixels at the same position in the background image.  

For example, if a pixel at ($x_0$, $y_0$) is an artifact candidate, the replacement process is as follows. For each pixel at ($x_0$, $y_0$) and its neighbours, the absolute difference $D^i(x_n,y_n)$ between its value in the background image and each input image are computed as follows respectively,  
 \begin{equation}
 D^i(x_n,y_n)=|\boldsymbol{I}^i{(x_n,y_n)}-\boldsymbol{I}_{\text{back}}(x_n,y_n)|, \text{ for }{(x_n,y_n)\in \mathcal{N}},
\label{eqn:bgdiff}
\end{equation}
where $ i \in \{1,...,k\}$, and $\mathcal{N}$ is a set of neighbor pixels of the artifact candidate pixel at ${(x_0,y_0)}$.

If the difference $D^i(x_n,y_n)$ is smaller than a predefined threshold $\tau_4$, the pixels at ($x_n$, $y_n$) in the background image and input image are considered the same, and the thresholded difference ${D_{\text {t}}}^i(x_n,y_n)$ is set to a certain value. In this work, it is set to 1 for simplicity,

\begin{equation}
{D_{\text {t}}}^i{(x_n,y_n)}=\left\lbrace
\begin{array}{ll}
1, & \text{ if } D^i(x_n,y_n)\leq  \tau_4,\\
0, & \text{ if } D^i(x_n,y_n)> \tau_4,
\end{array},
\text{ for }{(x_n,y_n)\in \mathcal{N}}.
\right.
\end{equation}

A Gaussian weighted sum of the thresholded differences ${D_{\text{w}}}^i{(x_0,y_0)}$ for each input image is computed as follows respectively,
\begin{equation}
{D_{\text{w}}}^i{(x_0,y_0)}=\sum_{(x_n,y_n)\in \mathcal{N}} {A*\exp{(-{(\frac{(x_n-x_0)^2}{2\sigma^2_x}+\frac{(y_n-y_0)^2}{2\sigma^2_y})   })}}*{D_{\text {t}}}^i{(x_n,y_n)},
\label{eqn:GaussianWeighted}
\end{equation}

A larger Gaussian weighted sum ${D_{\text{w}}}^i{(x_0,y_0)}$ stands for that the $i^{\text{th}}$ image is closer to the background image at ($x_0$, $y_0$).
If ${D_{\text{w}}}^{i_0}{(x_0,y_0)}$ is the largest weighted sum, and the absolute difference between the pixel at ($x_0$, $y_0$) in the background image $\boldsymbol{I}_{\text{back}}(x_0,y_0) $ and the corresponding pixel in $i^{\text{th}}_0$ input image $\boldsymbol{I}^{i_0}{(x_0,y_0)}$ is above a predefined threshold $\tau_5$, 
 $\boldsymbol{I}_{\text{back}}(x_0,y_0) $ is replaced by $\boldsymbol{I}^{i_0}{(x_0,y_0)}$. The replacement is as follows,
\begin{equation}
i_0 = \arg\max_i{\{...,{D_{\text{w}}}^i{(x_0,y_0)},...\}}, \text{ for } i_0,i \in\{1,...,k\} ,
\end{equation}
\begin{equation}
\boldsymbol{I}_{\text{back}}(x_0,y_0)=\boldsymbol{I}^{i_0}{(x_0,y_0)}, 
\text{ if } |\boldsymbol{I}_{\text{back}}(x_0,y_0)-\boldsymbol{I}^{i_0}{(x_0,y_0)}|>\tau_5.
\label{eqn:tau5}
\end{equation}

If a pixel in the background image is replaced, its neighbours and itself will be considered as artifact candidates in the next iteration. The artifact removal process stops until the background converges.

\subsection{Experimental setup}

\begin{figure}[tbh]
\centering

\vspace{3pt}

\begin{minipage}[b]{0.19\linewidth}
\subfigure[${image 01}$]{
\includegraphics[width=\textwidth]{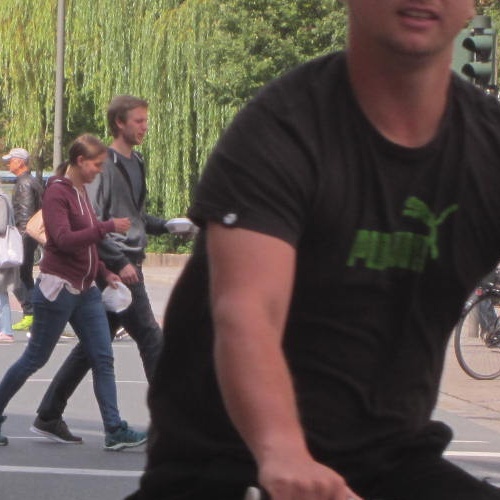}
\label{subfig:RawImage0}

}
\end{minipage}
\begin{minipage}[b]{0.19\linewidth}
\subfigure[$image 02$]{
\includegraphics[width=\textwidth]{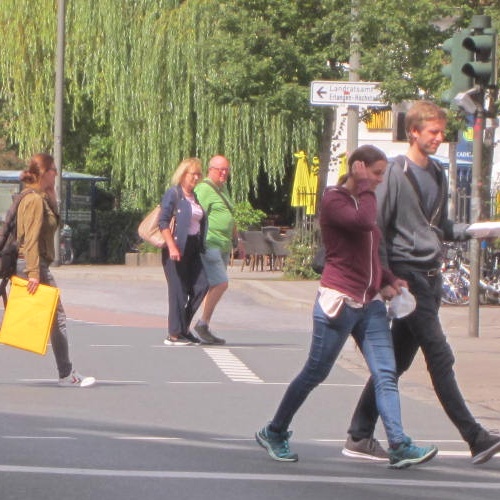}
\label{subfig:RawImage1}
}
\end{minipage}
\begin{minipage}[b]{0.19\linewidth}
\subfigure[$image 03$]{
\includegraphics[width=\textwidth]{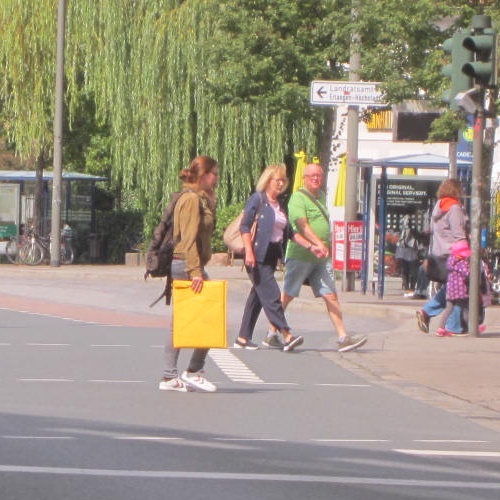}
\label{subfig:RawImage2}
}
\end{minipage}
\begin{minipage}[b]{0.19\linewidth}
\subfigure[$image 04$]{
\includegraphics[width=\textwidth]{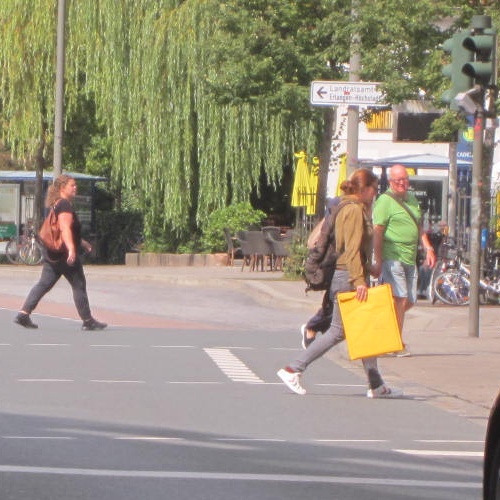}
\label{subfig:RawImage3}
}
\end{minipage}
\begin{minipage}[b]{0.19\linewidth}
\subfigure[$image 05$]{
\includegraphics[width=\textwidth]{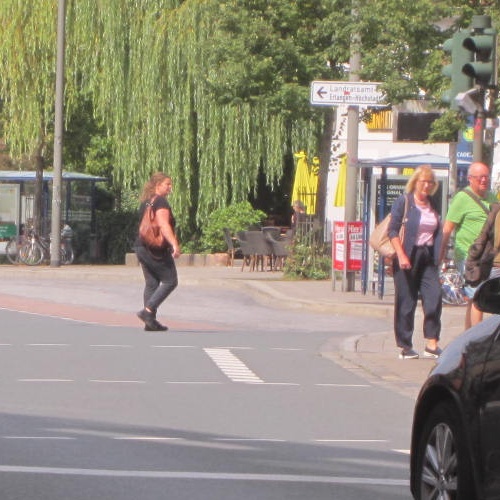}
\label{subfig:RawImage4}
}
\end{minipage}
\begin{minipage}[b]{0.19\linewidth}
\subfigure[$image 06$]{
\includegraphics[width=\textwidth]{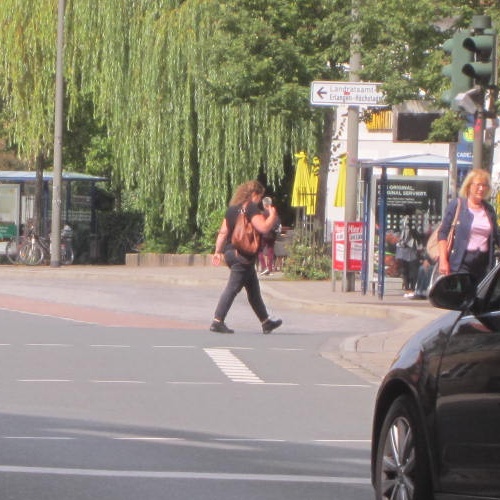}
\label{subfig:RawImage5}
}
\end{minipage}
\begin{minipage}[b]{0.19\linewidth}
\subfigure[ $image 07$]{
\includegraphics[width=\textwidth]{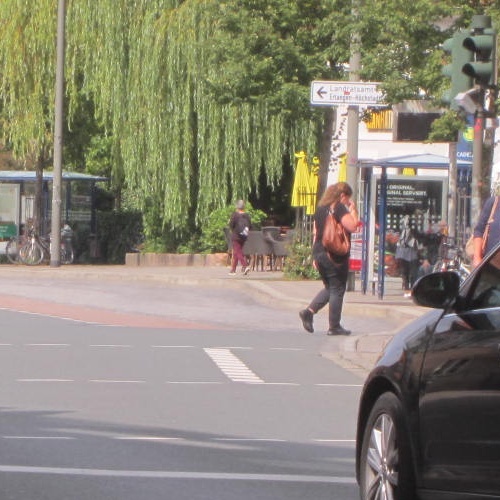}
\label{subfig:RawImage6}
}
\end{minipage}
\begin{minipage}[b]{0.19\linewidth}
\subfigure[$image 08$]{
\includegraphics[width=\textwidth]{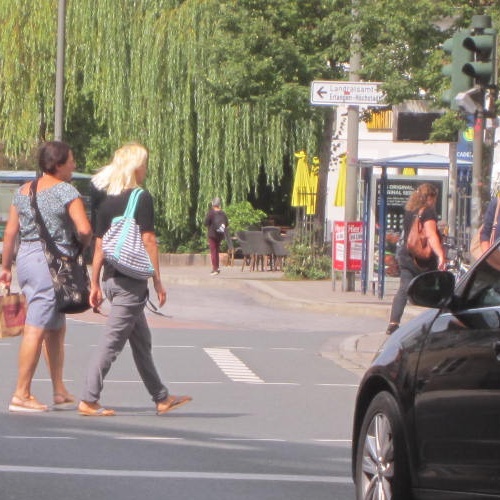}
\label{subfig:RawImage7}
}
\end{minipage}
\begin{minipage}[b]{0.19\linewidth}
\subfigure[$image 09$]{
\includegraphics[width=\textwidth]{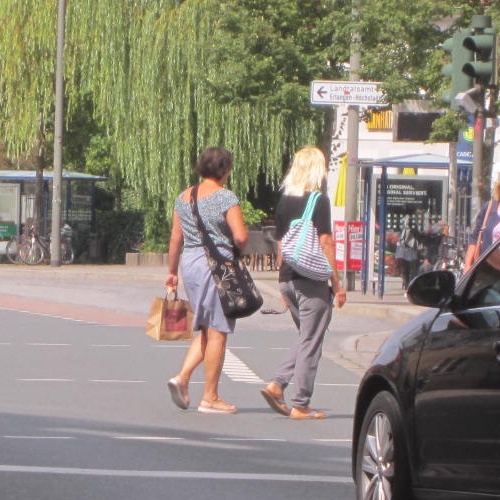}
\label{subfig:RawImage8}
}
\end{minipage}
\begin{minipage}[b]{0.19\linewidth}
\subfigure[$image 10$]{
\includegraphics[width=\textwidth]{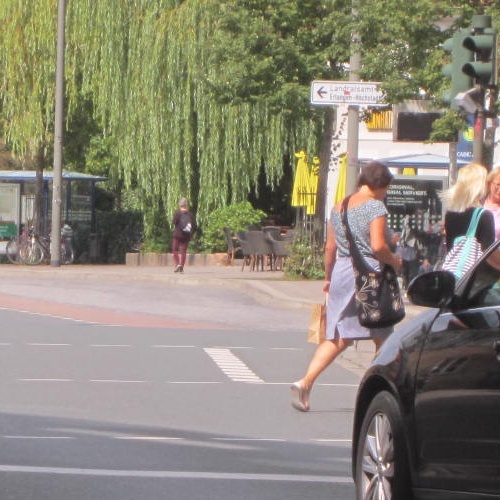}
\label{subfig:RawImage9}
}
\end{minipage}

\caption{ Ten images from the custom $Arcaden$ dataset. }
\label{fig:Arcaden dataset}

\end{figure}

In order to verify the validity of the  proposed background recovery method, an experiment on a outdoor dataset is carried out.  A set of 10 images (size 500*500, 24-bit color) named $Arcaden$, taken by a static camera in burst mode, are shown in Fig.~\ref{fig:Arcaden dataset}. The C++ with OpenCV implementation takes half an hour on a computer with 2.8 GHz i7-7700 CPU and 16 Gb of RAM .

The threshold ($\tau_0$) in Eqn.~(\ref{eqn:seeds})  for computing the background seeds is crucial for determining background seeds. If $\tau_0$ is too large, some foreground pixels may be mislabelled as background seeds. If it is too small, there may be not enough seeds. In this work, $\tau_0$ is set to 10. 

Superpixels are produced with \textit{createSuperpixelSLIC} function in OpenCV. In this function, five parameters are set as follows:
		 \textit{algorithm} = 0,
		 \textit{region\_size} = 15,
		 \textit{ruler} = 15,
		 \textit{num\_iterations} = 10.
There are three optional algorithms, SLIC, SLICO and MSLIC, to choose. Here SLIC is chosen. The \textit{region\_size} parameter is an average superpixel size. It should be set carefully, because background objects may not be segmented correctly if \textit{region\_size} is too large, and computation cost is high if region size is too small. \textit{ruler} is a superpixel smoothness parameter. \textit{num\_iterations} is the number of iterations. 

If the number of pixels labelled as background in a superpixel is over $\tau_1$ in Eqn.~(\ref{eqn:tau1}) , which is set to 25, or the ratio between the number of labelled pixels and the number of pixels in the superpixel is over $\tau_2$ in Eqn.~(\ref{eqn:tau1}) , which is set to 0.4, all the pixels inside the superpixel are labelled as background.  

$\tau_3$ in Eqn.~(\ref{eqn:tau3}) is set to 10 to update the candidate masks. When it iterates over 100 times or less than three pixels in the background image change after an iteration, the iteration terminates. The neighbourhood size is $41\times 41$ for $\mathcal{N}$ in Eqn.~(\ref{eqn:bgdiff}). $\tau_4$ is set to 10. The variance parameter $\sigma_x$ and $\sigma_y$ for the 2-D Gaussian weighted function are set to 20, and the kernel size is also $41\times 41$. $\tau_5$ in Eqn.~(\ref{eqn:tau5}) is set to 5.

\begin{figure}
\centering
\begin{minipage}[b]{0.3\linewidth}
\subfigure[Background seeds]{
\includegraphics[width=\textwidth]{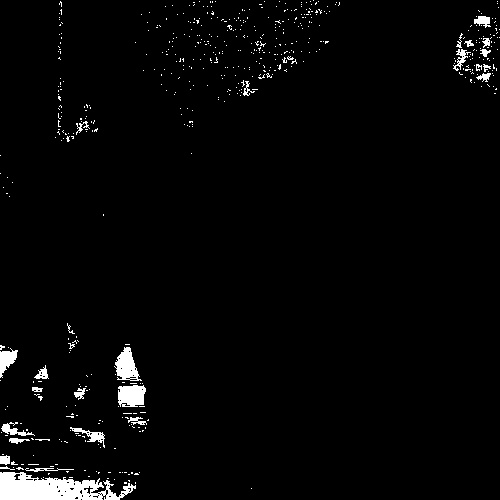}
\label{subfig:ImSeedsname}
}
\end{minipage}
\begin{minipage}[b]{0.3\linewidth}
\subfigure[Superpixels]{
\includegraphics[width=\textwidth]{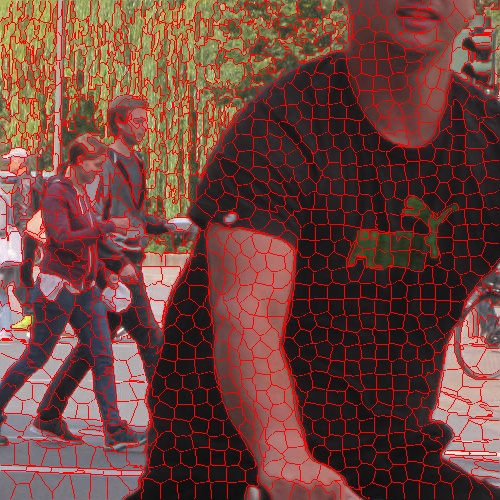}
\label{subfig:result0}
}
\end{minipage}
\begin{minipage}[b]{0.3\linewidth}
\subfigure[Initialized candidate mask]{
\includegraphics[width=\textwidth]{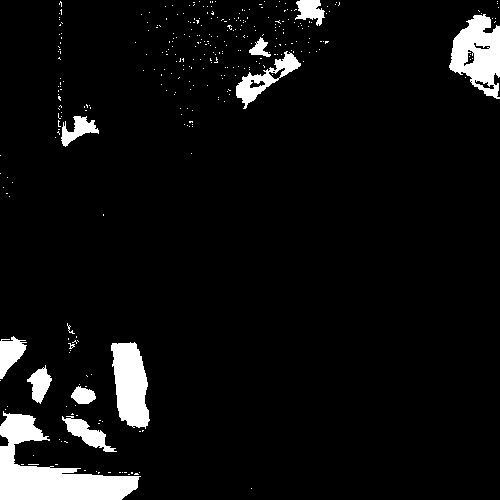}
\label{subfig:Image0Iteration0templateM}
}
\end{minipage}

\begin{minipage}[b]{0.3\linewidth}
\subfigure[Initialized candidate image]{
\includegraphics[width=\textwidth]{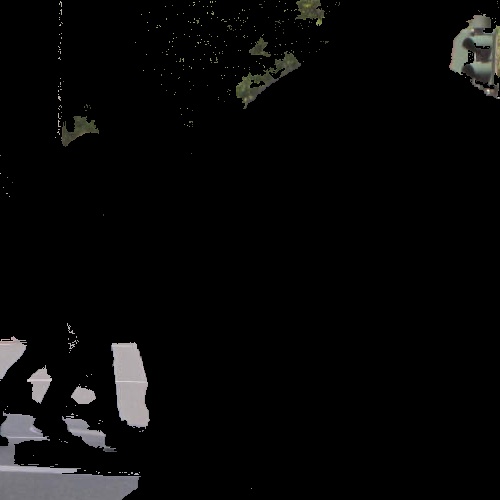}
\label{subfig:Image0Iteration0raw}
}
\end{minipage}
\begin{minipage}[b]{0.3\linewidth}
\subfigure[Initialized background]{
\includegraphics[width=\textwidth]{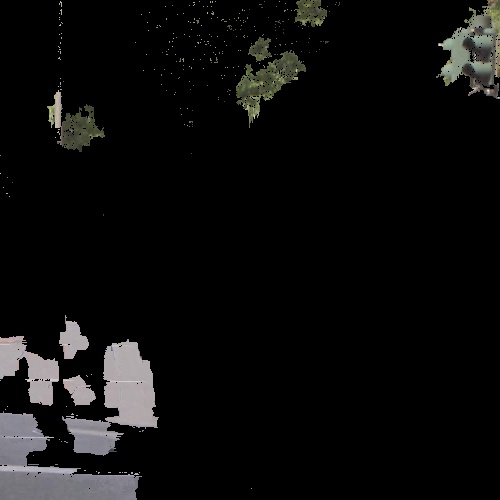}
\label{subfig:RGBAfterIteration0algorithm0region_size15ThresholdRatio0400000ThresholdPixels25}
}
\end{minipage}
\begin{minipage}[b]{0.3\linewidth}
\subfigure[Final candidate mask]{
\includegraphics[width=\textwidth]{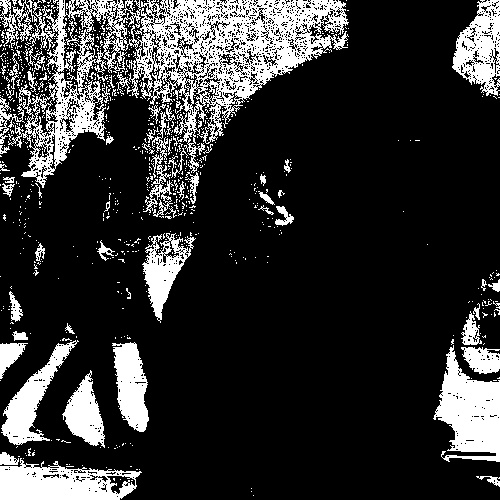}
\label{subfig:Image0Iteration37templateM}
}
\end{minipage}

\begin{minipage}[b]{0.3\linewidth}
\subfigure[Final candidate image]{
\includegraphics[width=\textwidth]{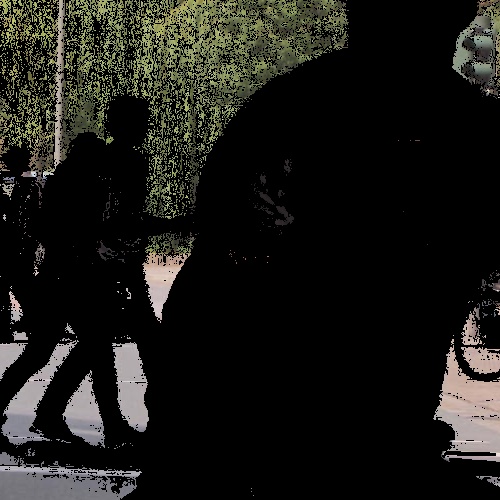}
\label{subfig:Image0Iteration37raw}
}
\end{minipage}
\begin{minipage}[b]{0.3\linewidth}
\subfigure[Background image with ghosting artifacts]{
\includegraphics[width=\textwidth]{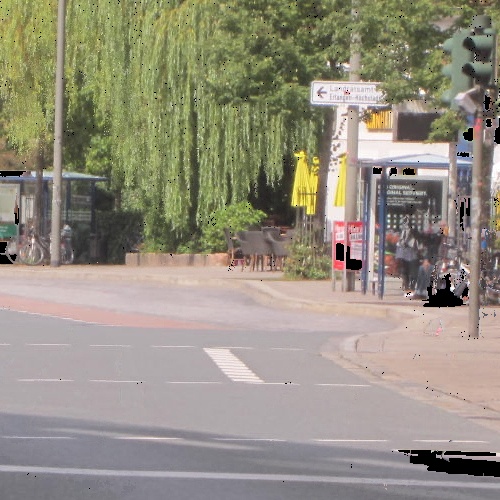}
\label{subfig:RGBAfterIteration37algorithm0region_size15ThresholdRatio0400000ThresholdPixels25}
}
\end{minipage}
\begin{minipage}[b]{0.3\linewidth}
\subfigure[Output]{
\includegraphics[width=\textwidth]{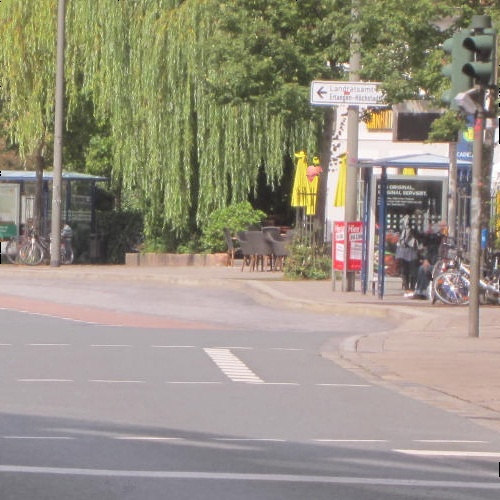}
\label{subfig:imAfterEdgeRGBname56}
}
\end{minipage}
\caption{ (a) Background seeds, (b) superpixels of $image 01$ segmented by SLIC, (c) initialized background candidate mask of $image 01$, (d) initialized background candidate image of $image 01$, (e) initialized background image, (f) final background candidate mask of $image 01$, (g) final background candidate image of $image 01$, (h) background image of the last iteration with ghosting artifacts, and (i) output background after ghosting artifact removal.}
\label{fig:result}
\end{figure}

\section{Results}

The results are displayed in Fig.~\ref{fig:result}.  The background seeds based on Eqn.~(\ref{eqn:seeds}) are shown in Fig.~\ref{subfig:ImSeedsname}. Because of the strict constraint, many background pixels are not considered as backgroud seeds.  A segmentation result of $image 01$ by SLIC is displayed in Fig.~\ref{subfig:result0}.

Background candidate  masks are then initialized. The initialized mask of $image 01$ is presented in Fig.~\ref{subfig:Image0Iteration0templateM}. Compared to Fig.~\ref{subfig:ImSeedsname}, the area of background has grown.
By masking out each input image with corresponding mask, background candidate images are initialized. The result of $image 01$  is shown in Fig.~\ref{subfig:Image0Iteration0raw}. 
Afterwards the background image is initialized. It is shown in Fig.~\ref{subfig:RGBAfterIteration0algorithm0region_size15ThresholdRatio0400000ThresholdPixels25}). The area of background is larger than that in Fig.~\ref{subfig:Image0Iteration0raw}.

The final candidate mask of $image 01$ is shown in Fig.~\ref{subfig:Image0Iteration37templateM}. There are some pixels misclassified, due to the similarity between the background pixels and foreground pixels. The final background candidate image of  $image 01$ can be found in Fig.~\ref{subfig:Image0Iteration37raw}. It shows that the foreground objects are omitted. As shown in Fig.~\ref{subfig:RGBAfterIteration37algorithm0region_size15ThresholdRatio0400000ThresholdPixels25}, the background updated has ghosting artifacts. The output background after artifact removal is displayed in Fig.~\ref{subfig:imAfterEdgeRGBname56}. It shows that the background image is recovered successfully.

\section{Discussion and Conclusion}

In this work, we proposed an intuitive background recovery method based on superpixels. The algorithm starts with seeking only background seeds. Then background candidate  masks are initialized with the seeds. Afterwards mask out input images with the  masks to create background candidate images. The background image is built by choosing pixels from those images and then updated until converge. Finally ghosting artifacts are removed by the k-NN method. 

The proposed method requires three assumptions. Firstly, enough background pixels need to exist for background seeds.
Secondly, the background should be relatively stationary, e.g., waving trees in a small local region is acceptable.
Finally, each region of the background need to show in several frames.

The most prominent advantage of this method is that it can recover a background image from a heavily occluded dataset, because pixels in the background image are chosen from background candidate images, which ``grow" from the background seeds. The background seeds absolutely belong to background, no matter how frequently foreground objects appear spatially or temporally in the dataset. If each region of the background shows in several frames, the two update rules for background candidate masks will work and a clean background can be recovered.

We have demonstrated our algorithm on the custom $Arcaden$ dataset. Though there are some severely occluded regions, it has removed all moving objects, although a car is stationary in the last few frames.

 However, some limitations exist for the proposed algorithm. When transient object shadows are involved, the foreground is difficult to segment. Or when foreground objects and background share the same color at a same position, the proposed algorithm may have poor performance, because of the first update rule for masks.

%

\begin{thebibliography}{10}
\providecommand{\url}[1]{#1}
\csname url@samestyle\endcsname
\providecommand{\newblock}{\relax}
\providecommand{\bibinfo}[2]{#2}
\providecommand{\BIBentrySTDinterwordspacing}{\spaceskip=0pt\relax}
\providecommand{\BIBentryALTinterwordstretchfactor}{4}
\providecommand{\BIBentryALTinterwordspacing}{\spaceskip=\fontdimen2\font plus
\BIBentryALTinterwordstretchfactor\fontdimen3\font minus
  \fontdimen4\font\relax}
\providecommand{\BIBforeignlanguage}[2]{{%
\expandafter\ifx\csname l@#1\endcsname\relax
\typeout{** WARNING: IEEEtran.bst: No hyphenation pattern has been}%
\typeout{** loaded for the language `#1'. Using the pattern for}%
\typeout{** the default language instead.}%
\else
\language=\csname l@#1\endcsname
\fi
#2}}
\providecommand{\BIBdecl}{\relax}
\BIBdecl

\bibitem{BOUWMANS20173}
T.~Bouwmans, L.~Maddalena, and A.~Petrosino, ``Scene background initialization:
  A taxonomy,'' \emph{Pattern Recognition Letters}, vol.~96, pp. 3 -- 11, 2017,
  scene Background Modeling and Initialization.

\bibitem{TemporalMedian}
M.~{Hung}, J.~{Pan}, and C.~{Hsieh}, ``Speed up temporal median filter for
  background subtraction,'' pp. 297--300, Sep. 2010.

\bibitem{784637}
C.~{Stauffer} and W.~E.~L. {Grimson}, ``Adaptive background mixture models for
  real-time tracking,'' in \emph{Proceedings. 1999 IEEE Computer Society
  Conference on Computer Vision and Pattern Recognition (Cat. No PR00149)},
  vol.~2, June 1999, pp. 246--252 Vol. 2.

\bibitem{s120912279}
J.~Lee and M.~Park, ``An adaptive background subtraction method based on kernel
  density estimation,'' \emph{Sensors}, vol.~12, no.~9, pp. 12\,279--12\,300,
  2012.

\bibitem{NovelRobustStatistica}
H.~Wang and D.~Suter, ``A novel robust statistical method for background
  initialization and visual surveillance,'' 01 2006, pp. 328--337.

\bibitem{Automaticocclusion}
C.~{Herley}, ``Automatic occlusion removal from minimum number of images,'' in
  \emph{IEEE International Conference on Image Processing 2005}, vol.~2, Sep.
  2005, pp. II--1046.

\bibitem{spatiotemporalcontinuities}
S.~{Varadarajan}, L.~J. {Karam}, and D.~{Florencio}, ``Background subtraction
  using spatio-temporal continuities,'' in \emph{2010 2nd European Workshop on
  Visual Information Processing (EUVIP)}, July 2010, pp. 144--148.

\bibitem{PatchBasedBackground}
A.~{Colombari} and A.~{Fusiello}, ``Patch-based background initialization in
  heavily cluttered video,'' \emph{IEEE Transactions on Image Processing},
  vol.~19, no.~4, pp. 926--933, April 2010.

\bibitem{Backgroundrecovery}
A.~{Shrotre} and L.~J. {Karam}, ``Background recovery from multiple images,''
  in \emph{2013 IEEE Digital Signal Processing and Signal Processing Education
  Meeting (DSP/SPE)}, Aug 2013, pp. 135--140.

\bibitem{SuperpixelsMotion}
J.~Lim and B.~Han, ``Generalized background subtraction using superpixels with
  label integrated motion estimation,'' in \emph{Computer Vision -- ECCV 2014},
  D.~Fleet, T.~Pajdla, B.~Schiele, and T.~Tuytelaars, Eds.\hskip 1em plus 0.5em
  minus 0.4em\relax Cham: Springer International Publishing, 2014, pp.
  173--187.

\bibitem{SuperpixelsSimilarity}
W.~{Fang}, T.~{Zhang}, C.~{Zhao}, D.~B. {Soomro}, R.~{Taj}, and H.~{Hu},
  ``Background subtraction based on random superpixels under multiple scales
  for video analytics,'' \emph{IEEE Access}, vol.~6, pp. 33\,376--33\,386,
  2018.

\bibitem{Extractingessence}
M.~Alexa, ``Extracting the essence from sets of images,'' \emph{Proc. of the
  Eurographics Workshop on Computational Aesthetics}, pp. 113--120, 01 2007.

\bibitem{RandomWalk}
K.~{Hua}, H.~{Wang}, C.~{Yeh}, W.~{Cheng}, and Y.~{Lai}, ``Background
  extraction using random walk image fusion,'' \emph{IEEE Transactions on
  Cybernetics}, vol.~48, no.~1, pp. 423--435, Jan 2018.

\bibitem{superpixelsstateoftheart}
D.~Stutz, A.~Hermans, and B.~Leibe, ``Superpixels: An evaluation of the
  state-of-the-art,'' \emph{Computer Vision and Image Understanding}, vol. 166,
  pp. 1--27, 2018.

\bibitem{guyon2012robust}
C.~Guyon, T.~Bouwmans, E.-h. Zahzah \emph{et~al.}, ``Robust principal component
  analysis for background subtraction: Systematic evaluation and comparative
  analysis,'' \emph{Principal component analysis}, vol.~10, pp. 223--238, 2012.

\bibitem{SuperpixelMatrix}
S.~{Javed}, S.~H. {Oh}, A.~{Sobral}, T.~{Bouwmans}, and S.~K. {Jung},
  ``Background subtraction via superpixel-based online matrix decomposition
  with structured foreground constraints,'' in \emph{2015 IEEE International
  Conference on Computer Vision Workshop (ICCVW)}, Dec 2015, pp. 930--938.

\bibitem{SLIC}
R.~Achanta, A.~Shaji, K.~Smith, A.~Lucchi, P.~Fua, and S.~Süsstrunk, ``Slic
  superpixels,'' p.~15, 2010.

\bibitem{meanshift}
D.~{Comaniciu} and P.~{Meer}, ``Mean shift: a robust approach toward feature
  space analysis,'' \emph{IEEE Transactions on Pattern Analysis and Machine
  Intelligence}, vol.~24, no.~5, pp. 603--619, May 2002.

\end{thebibliography}



\end{document}